\def\year{2022}\relax
\documentclass[letterpaper]{article} 
\usepackage{aaai22}  
\usepackage{times}  
\usepackage{helvet}  
\usepackage{courier}  
\usepackage[hyphens]{url}  
\usepackage{graphicx} 
\usepackage{natbib}  
\usepackage{caption} 
\DeclareCaptionStyle{ruled}{labelfont=normalfont,labelsep=colon,strut=off} 
\usepackage{multirow}
\usepackage{amsfonts}
\usepackage{tabularx}
\usepackage{booktabs}
\usepackage{amsmath}
\usepackage{enumitem} 
\DeclareMathOperator*{\argmax}{arg\,max}
\usepackage{xcolor}

\newcommand{\ours}{\textbf{SRIC}}

\title{Anti-Asian Hate Speech Detection via Data Augmented Semantic Relation Inference}
\author{
   Jiaxuan Li and Yue Ning\\
}
\affiliations{
   Stevens Institute of Technology\\
     \{jli237, yue.ning\}@stevens.edu
}

\begin{document}

\maketitle

\begin{abstract}
With the spreading of hate speech on social media in recent years, automatic detection of hate speech is becoming a crucial task and has attracted attention from various communities. This task aims to recognize online posts (e.g., tweets) that contain hateful information. 
The peculiarities of languages in social media, such as short and poorly written content, lead to the difficulty of learning semantics and capturing discriminative features of hate speech. 
Previous studies have utilized additional useful resources, such as sentiment hashtags, to improve the performance of hate speech detection. 
Hashtags are added as input features serving either as sentiment-lexicons or extra context information. 
However, our close investigation shows that directly leveraging these features without considering their context may introduce noise to classifiers. 
In this paper, we propose a novel approach to leverage sentiment hashtags to enhance hate speech detection in a natural language inference framework. We design a novel framework \textbf{SRIC} that simultaneously performs two tasks: (1) \textbf{s}emantic \textbf{r}elation \textbf{i}nference between online posts and sentiment hashtags, and (2) sentiment \textbf{c}lassification on these posts. The semantic relation inference aims to encourage the model to encode sentiment-indicative information into representations of online posts. We conduct extensive experiments on two real-world datasets and demonstrate the effectiveness of our proposed framework compared with state-of-the-art representation learning models.

\end{abstract}

\section{Introduction}
Hate speech is spreading rapidly and widely on social media at an unprecedented rate, which has resulted in a large number of victims and unhealthy online environments.
For example, during the outbreak of COVID-19, racism and hateful topics against Asian groups were rampant. The widespread of online hate speeches ultimately leads to a surge in real-life hate crimes. 
However, automatically detecting online hate speech faces several challenges.
First, it is difficult to track and collect related data from massive posts in social media to train hate speech classifiers.
Moreover, informal languages in social media, with the exception of linguistic complexity of language, make hate speech detection more difficult. It is necessary to recognize sentiment informative features that can help identify hate speech.
\begin{table}[t]
\centering
\begin{tabular}{lp{4cm}}
\toprule
     Sentiment & Posts  \\
\midrule
     Negative & Rainy days always make me feel sad \#ihaterain  \\
        \cmidrule{1-2}
   Positive & I prefer the fresh air on rainy days \#iredhaterain \\
\bottomrule
\end{tabular}
\caption{Example of Inconsistent Sentiment between Online Posts and their Hashtags}
\label{controversy_example}
\end{table}
In order to overcome these challenges, the extensive use of hashtags in social media has drawn growing attention from researchers ~\cite{davidov2010enhanced,wang2011topic,koto2015hbe,kouloumpis2011twitter}.
Hashtags are often added as metadata to a textual utterance with the goal of increasing visibility and speeding up dissemination. Typically, hashtags begin with a hash symbol, followed by a word or a phrase without separating blanks, such as \#racism or \#racismisvirus.
On the one hand, hashtags can serve as user-annotated topics, which are used as keywords to identify and track hot topics in explosive online information. Researchers can collect related data by searching hashtags that interest them. 
On the other hand, hashtags contain rich semantic information, which may indicate opinion tendencies from their literal information. 
For instance, hashtag \emph{\#iloveBiber} conveys the positive sentiment through the word \emph{love}, and \emph{\#ihaterain} expresses the negative sentiment through the word \emph{hate}.
Such sentiment hashtags that sufficiently indicate sentiment polarity have been utilized in many studies of sentiment classification. 
Rezapour \textit{et al.}~\cite{rezapour2017identifying} manually annotate sentiment hashtags in posts and add them to an existing sentiment lexicon. Afterwards, unsupervised lexicon-based sentiment analysis algorithms are applied to classify posts by analyzing the characteristics of words in the post, such as frequency and co-occurrence. Some researchers utilize the presence of sentiment hashtags to label posts and use supervised learning algorithms to train sentiment classifiers~\cite{davidov2010enhanced,kouloumpis2011twitter}. Mounica Maddela \textit{et al.}~\cite{maddela2019multi} use sentiment hashtags as complementary information to enrich the context of tweets.
However, the previous literature only focuses on the semantics of sentiment hashtags while ignoring their context which may introduce noisy into sentiment classifiers.
As shown in Table~\ref{controversy_example}, 
the first example has consistent sentiment between the post and the hashtag, but the post in the second example shows the opposite sentiment with the hashtag. Therefore, directly incorporating sentiment hashtags without considering their context will confuse classifiers by introducing inconsistent sentiment signals. This observation calls for further investigating the effectiveness of sentiment hashtags and the way to combine their semantics in sentiment classification.

In this paper, we propose a novel approach that leverages sentiment hashtags to enhance hate speech detection on multi-class classification tasks, instead of binary classifications (e.g., positive or negative). 
We first extract sentiment hashtags from posts and convert them into a word sequence by a word segmentation tool. For example, \#racismisvirus is converted to ``racism is virus''. 
By converting sentiment hashtags into meaningful sequences of words, the semantic information is fully exposed.
Next, instead of directly adding them back into posts to expand textual context, we propose \ours, a \textbf{S}emantic \textbf{R}elation \textbf{I}nference model for \textbf{C}lassifying hate-related online posts while capturing the semantic relations between the posts and these hashtags.
Through incorporating such semantic relational knowledge into hate speech classifiers, we expect to embed fine-grained sentiment information into the representations of words to improve hate speech detection performance.
The main contributions of this work are summarized as follows:

\begin{itemize}[leftmargin=*] 
\item We investigate an effective way to leverage sentiment hashtags for hate speech detection problems. 
Instead of directly incorporating sentiment hashtags as features without considering their context, 
we propose a novel framework to learn the semantic relations between online posts and sentiment hashtags and further incorporate these relational information in the classification task (i.e., hate, counter-hate, or neutral). Through semantic relation modeling, the framework is able to learn consistent sentiment features and mitigate the situation where inconsistent sentiment signals exist in posts.

\item We propose a data-augmented extension to our framework to utilize data samples that do not contain hashtags. 
The data augmentation technique provide pseudo-hashtags for posts and can be applied to other frameworks as well.
The sentiment classification ability of \ours~is further improved via training on such augmented sample pairs.

\item We conduct extensive experiments and illustrate the proposed model's effectiveness on two multi-class datasets. The experimental results show that sentiment hashtags can be exploited as valuable information to improve the accuracy of hate speech detection. Moreover, the results demonstrate that our proposed approach with augmented sentiment hashtags is effective to capture discriminative semantic features for classifying sentiments.
\end{itemize}

\section{Related Work}
Automatic hate speech detection in social media remains an essential task that has not yet been fully addressed. 
Hate speech includes abusive and aggressive languages to attack individuals or groups~\cite{schmidt2017survey,zhang2019hate,fortuna2018survey}. A large number of studies have been conducted to identify different types of hate speeches based on race, color, ethnicity, gender, sexual orientation, nationality, and religion~\cite{kwok2013locate,davidson2019racial,saha2018hateminers,cowan2005heterosexuals,sap2019risk}.
Besides, there are also many case studies of hate speech on world topics such as immigrants~\cite{capozzi2019computational,indurthi2019fermi}, refugees~\cite{vazquez2019hate,frias2019hate}, and presidential elections~\cite{DBLP:conf/wassa/GrimmingerK21,siegel2021trumping}.
Recently, anti-Asian hate speeches have also received a lot of attention due to the outbreak of COVID-19 ~\cite{DBLP:conf/emnlp/HardageN20,fan2020stigmatization,DBLP:conf/icmla/VishwamitraH0CC20}. 
As a text classification task, the most important factor in hate speech detection is the construction of effective features. There are several types of features used in hate speech detection.
Surface features, such as Bag of Words (BOW), term frequency-inverse document frequency (TF-IDF), word and character n-grams, URL mentions, emojis, and hashtags, have been utilized as fundamental features~\cite{xia-etal-2020-demoting,DBLP:conf/icwsm/DavidsonWMW17,DBLP:conf/naacl/WaseemH16}. Lexical features are obtained by looking up specific abusive and offensive words in a specific sentiment lexicon and counting their frequencies~\cite{DBLP:journals/corr/abs-2104-12265,DBLP:conf/icwsm/DavidsonWMW17}. Linguistic features include syntactic information such as Part of Speech (PoS), dependency relations, and Named Entity Recognition (NER)~~\cite{DBLP:conf/ijcai/ZhongLSRG0C16,burnap2015cyber,DBLP:conf/ahfe/EnglmeierM20,DBLP:conf/acl-alw/NarangB20,DBLP:conf/acl-socialnlp/SchmidtW17}. Semantic features and embeddings identify the sense of words in the context of a sentence.
In recent years, deep learning has emerged as a powerful technique that learns hidden representations of data and achieved state-of-the-art (SOTA) prediction results on several NLP tasks. 
Deep learning based hate speech detection uses contextual word embeddings that are pre-trained on a large number of unlabeled corpora to encode semantic and syntactic features~\cite{DBLP:conf/www/BadjatiyaG0V17,DBLP:conf/icmla/MeltonBK20}. Other word embedding methods include fine-tuning using a labeled dataset and further constructing high-level intrinsic features through deep neural networks, such as Convolutional Neural Networks (CNNs)~\cite{DBLP:conf/coling/SantosG14, Gambck2017UsingCN,DBLP:conf/setn/GeorgakopoulosT18}, Recurrent Neural Networks (RNNs)~\cite{DBLP:conf/semeval/BaruahBD19,DBLP:conf/naacl/QianWEY21}, and transformers (e.g., BERT)~\cite{DBLP:conf/complexnetworks/MozafariFC19,DBLP:conf/icdm/SohnL19}.
Most existing work fails to handle the situation when inconsistent sentiment signals exist in a context.

\section{Dataset}
In this paper, we investigate two datasets: COVID-19 Anti-Asian dataset~~\cite{DBLP:journals/corr/abs-2005-12423} and East Asian Prejudice dataset~~\cite{DBLP:conf/acl-alw/VidgenHGMBWBHT20}. Labeled data are categorized into three sentiment categories: \textit{Anti-Asian hate}, \textit{Counter-hate}, and \textit{Neutral}. Anti-Asian hate tweets are speeches against Asian
groups. Counter hate tweets are speeches that condemn abuse against Asian groups or oppose racism. Neutral tweets are speeches that neither express hate nor counter-hate sentiment. 
Both datasets contain sentiment hashtags. For instance, the COVID-19 Anti-Asian dataset contains 11 hateful hashtags (e.g., \#chinavirus) and 5 counter-hate hashtags (e.g., \#racismisavirus) and East Asian Prejudice dataset contains 11 hateful hashtags.
\begin{figure}[t]
\centering
\includegraphics[width=1.0\columnwidth]{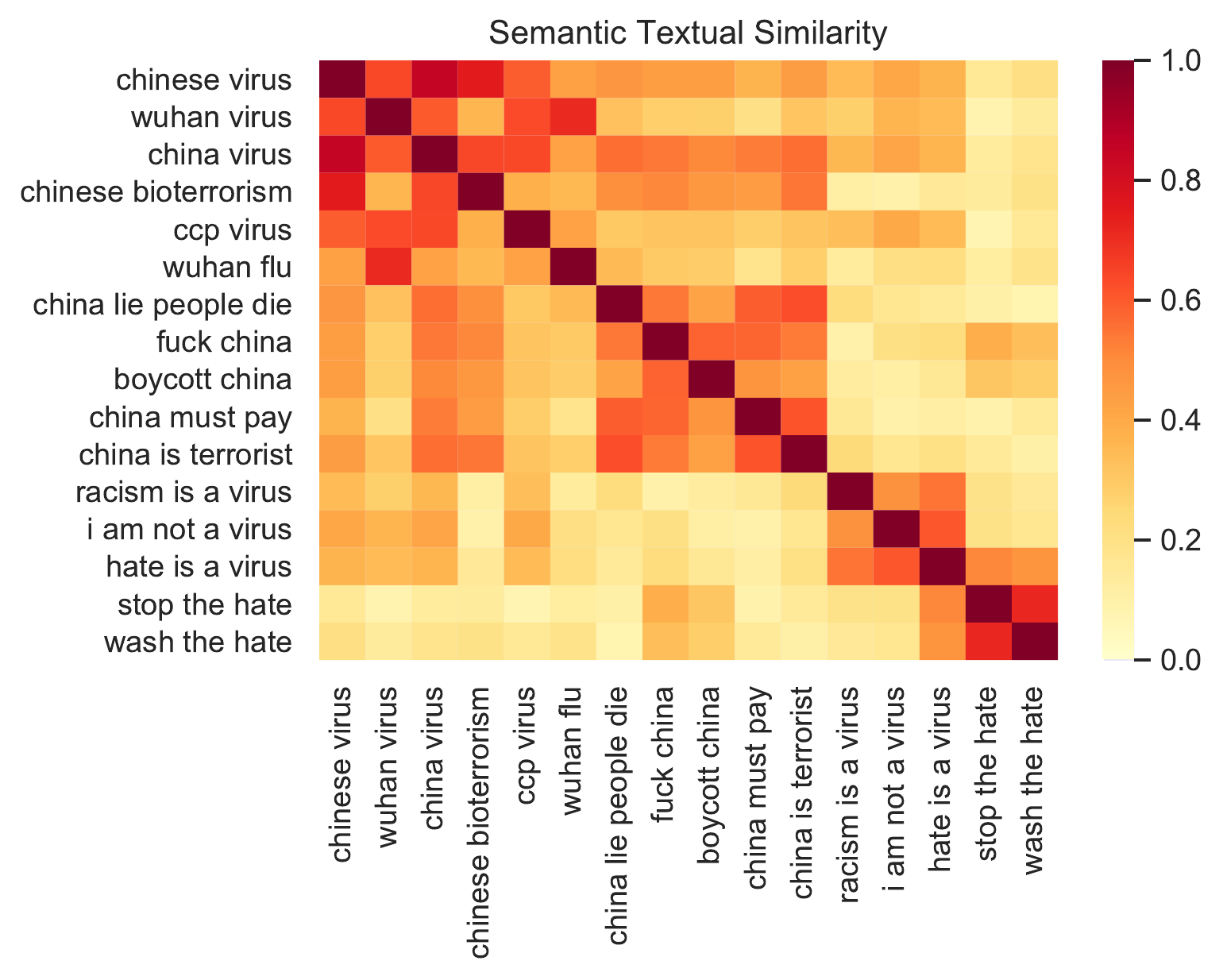} 
\caption{Semantic Similarity among Hashtags}
\label{simliarity}
\end{figure}
Figure \ref{simliarity} shows the semantic similarity of hateful and counter-hate hashtags in the COVID-19 Anti-Asian dataset. We can tell that there is a significant difference between the semantics of these two groups of hashtags. Such differences can help our model to capture discriminative sentiment features.

Most hashtags are attached or inserted to tweets for increasing visibility.
In some cases, they are very important for detecting the sentiment of tweets. It is difficult to discern the meaning of tweets solely based on their textual content~\cite{DBLP:conf/acl-alw/VidgenHGMBWBHT20}.
In other cases, they are less important to the meaning of tweets and their existence may introduce noisy sentiment signals. For example, hateful hashtags may appear in counter-hate or neutral tweets. 
As shown in Table \ref{tab:relation-example}, there are three types of relations between hateful hashtags and their context (i.e., tweets).  
In the first example, the tweet shows support of \#chinavirus. The post and the hashtag both express anti-Asian hateful emotions. In the second example, hashtag \#chinavirus has hateful sentiment, but the tweet itself exhibits counter-hate intentions. In the third example, the sentiment hashtag is only meant to indicate the topic. The content of the tweet neither supports nor opposes to \#chinavirus, which indicates neutral standpoints.
Therefore, directly incorporating these hashtags without considering their context will confuse hate speech classifiers by introducing inconsistent sentiment signals. 
\begin{table}[t]
\centering
\begin{tabular}{lp{4cm}}
\toprule
     Label & Post  \\
\midrule
   Hate& F**k china spreading the virus all over the world shame on you \#chinavirus \\
    \cmidrule{1-2}
   Counter-hate & Stop the xenophobia racism and hate \#chinavirus  \\
   \cmidrule{1-2}
   Neutral & \#Coronavirus report has been published first time and results are extremely serious. coronavirus has become the biggest challenge of in front of scientists and health experts \#chinavirus  \\
\bottomrule
\end{tabular}
\caption{Example of Different Sentiment Relations between Posts and Hashtags}
\label{tab:relation-example}
\end{table}

In this paper, we propose \ours~with the aim to detect different hate related labels while capturing the semantic relations between online posts and hashtags. The basis for determining the semantic relation between online posts and sentiment hashtags is whether they express consistent emotions. As show in Table~\ref{relation}, if posts and their included sentiment hashtags convey the same emotion, their semantic relation is entailment; otherwise, it is contradiction. For posts that do not convey specific emotions (hate-unrelated), their semantic relation with their sentiment hashtags should be neutral, which means that the relation is neither entailment nor contradiction. 
\begin{table}[t]
\begin{tabular}{m{1.5cm}ccc}
\toprule
  Hashtag Label &\multicolumn{3}{c}{Tweet Label}  \\
\cmidrule{2-4}
& Hate & Counter & Neutral\\
\cmidrule{2-4}
Hate& entailment & contradiction & neutral \\
 Counter & contradiction & entailment & neutral \\
\bottomrule
\end{tabular}
\caption{Inferred Relations between Posts and Hashtags. Counter means Counter-hate}
\label{relation}
\end{table}

\section{Methodology}

\subsection{Problem Formulation}
Suppose the training dataset is a collection of online posts $T = \{t_1, t_2,...,t_{N}\}$, where $N$ is the number of posts. A hate speech detection model $S(t)$ aims to predict the sentiment category of a post $t_i$: 
\begin{equation}
\hat{\mathbf{c}}_i = S(t_i),
\end{equation}
where $\hat{\mathbf{c}_i}$ is the predicted probability distribution over a set of pre-defined sentiment categories $C = \{C_1, C_2,...,C_M\}$ for post $t_i$, and $M$ is the number of sentiment labels. 
In this paper, we investigate three categories: \textit{hate}, \textit{counter-hate}, and \textit{neutral}. Specifically, we utilize sentiment hashtags and enhance the detection task by taking advantage of the semantic context of these hashtags. The segmented sentiment hashtags are denoted as $H = \{h_1, h_2, ..., h_K\}$, where $K$ is the number of sentiment hashtags.
According to their semantics, sentiment hashtags are manually categorized into two groups: hateful hashtags, and counter-hate hashtags.
Next, we extract the posts that contain at least one of the sentiment hashtags as $T_O$, and the remaining posts as $T_{O'}$.
For posts in $T_O$, we further split the content of each post into two parts, $T_O = \{(t_c, t_{h})_1,(t_c, t_{h})_2,...,(t_c, t_{h})_{|T_O|}\}$, where $t_{h}$ is one of the segmented sentiment hashtag in $H$ and $t_c$ is the content excluding $t_{h}$ in the tweet, and $|T_O|$ is the number of posts that contain sentiment hashtags. $T_{O'}$ denotes the set of tweets that do not contain sentiment hashtags, where $|T_{O'}|=N-|T_{O}|$.

A semantic relation inference model $I(t_c, t_{h})$ aims to predict the semantic relation between $t_c$ and $t_{h}$.
\begin{equation}
\hat{\mathbf{r}} = I(t_c, t_{h}),
\end{equation}
where $\hat{\mathbf{r}}$ is the predicted probability distribution over a set of semantic relations $R=\{R_1, R_2,...,R_J\}$ between the post $t_c$ and the hashtag $t_h$. In this paper, we investigate three semantic relations:~\textit{entailment}, \textit{contradiction}, and \textit{neutral}.
Next, we discuss how we design $S(\cdot)$ and $I(\cdot)$ in detail.

\subsection{The Proposed Model}
\begin{figure*}[t]
\centering
\includegraphics[width=1.0\textwidth]{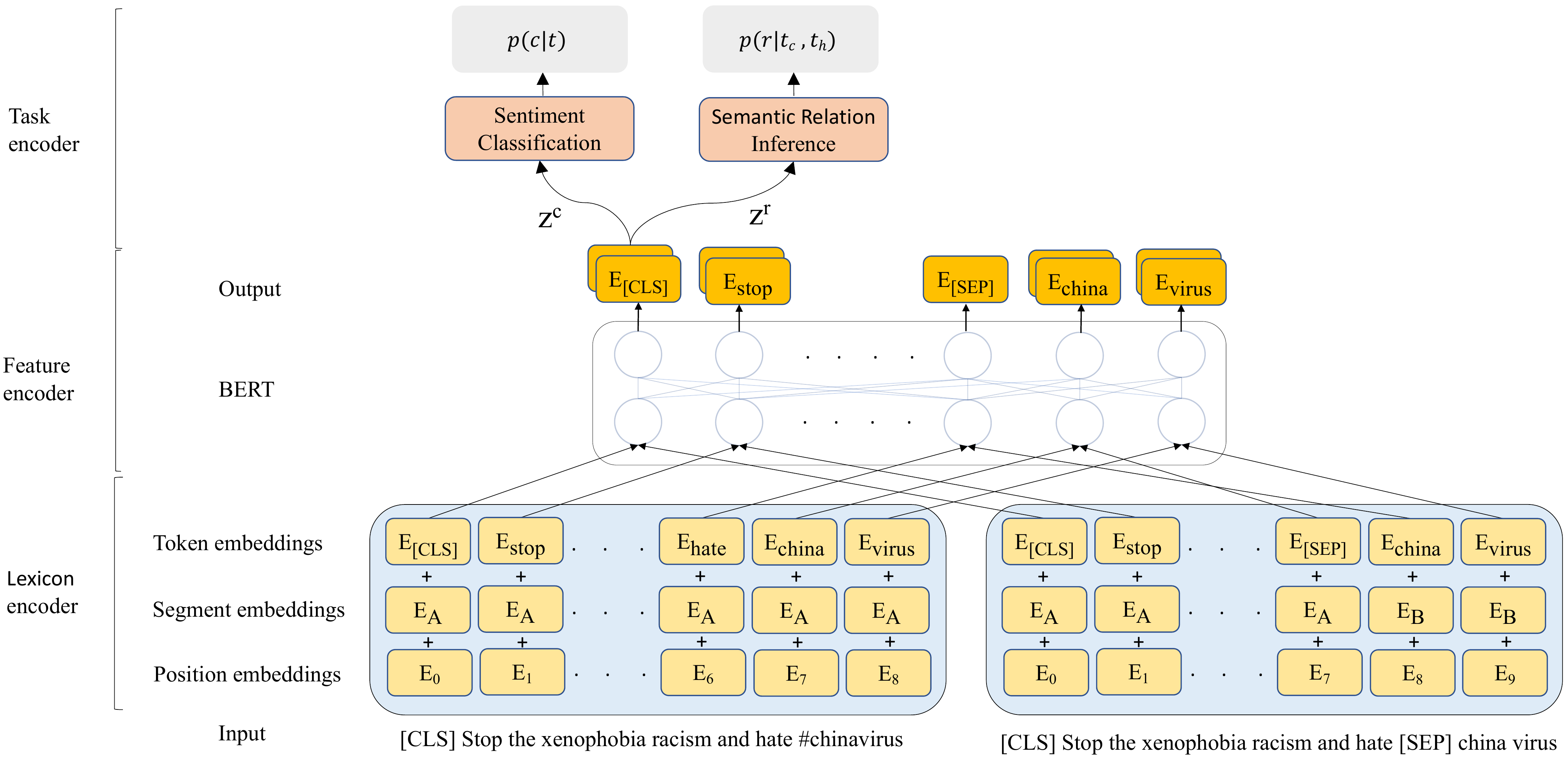} 
\caption{Overview of the proposed SRIC framework. It takes a post ($t_i$) and its corresponding pair $(t_c,t_h)_i$ as input. The lexicon encoder first converts a text input into a sequence of token representation vectors, where [SEP] token is a separator for the pair $(t_c,t_h)$. Next, these token vectors are fed to the BERT encoder to learn their contextualized embeddings during training. Finally, the hidden vector of [CLS] for the input $t_i$ is used for sentiment classification, while the hidden vector of [CLS] for the input $(t_c,t_h)$ is used for semantic relation inference.} 
\label{SRIC_framework}
\end{figure*}

In this section, we introduce our proposed framework \ours~based on one of the popular language models, BERT, as our backbone model. 
However, our method can be easily adapted to other pre-trained language models. An overview of the framework is shown in Figure \ref{SRIC_framework}. The framework consists of two components: 1) a semantic relation inference module, which aims to capture the semantic relation between posts and hashtags, and 2)  a classification module, which is designed to predict the sentiment categories of posts (i.e., \textit{hate}, \textit{counter-hate}, \textit{neutral}).

\subsubsection{Semantic Relation Inference.} Recent research has shown that paired text encoding is crucial in capturing complex relations between premise and hypothesis in natural language inference (NLI)~\cite{DBLP:conf/naacl/DevlinCLT19,DBLP:conf/emnlp/JiangM19}.  
We leverage the deep self-attention mechanism in BERT to encode an input sample as a pair: a post and its sentiment description (i.e., hashtags). Then  we extract the contextualized representation of this pair:
\begin{equation}
\mathbf{z}^r = \mbox{BERT}(t_c, t_{h}) \in \mathbb{R}^d,
\end{equation}
where $\mathbf{z}^r$ is a $d$-dimensional output vector corresponding to the special token $[\mbox{CLS}]$ as in the following input format: [$[\mbox{CLS}], t_c, [\mbox{SEP}], t_{h}$].

Next, we pass the resulting feature vector through a softmax activation layer to obtain the probability distribution of the semantic relations.

\begin{equation}
\hat{\mathbf{r}} = p(r|t_c,t_{h}) = \mbox{softmax}(W^r\mathbf{z}^r) \in \mathbb{R}^J,
\end{equation}
where $W^r\in \mathbb{R}^{J\times d}$ is a trainable weight matrix. 

This module is trained by using a cross-entropy loss with the ground-truth semantic relation labels of training data. The loss function is defined as below:
\begin{equation}
\mathcal{L}_{\mbox{infer}}=-\frac{1}{|T_O|}\sum_{i=1}^{T_O} \sum_{j=1}^{J} r_{i j} \log p\left(r_{ij} \mid {(t_c,t_h)_i}\right) ,
\end{equation}
where $r_{ij}$ denotes the ground truth of $i$-th sample on $j$-th class.
The inference module is expected to encode fine-grained sentiment information into features. 

Our assumption is that if a post entails a hashtag, the post and the hashtag should express similar semantics. Otherwise, they should have dissimilar semantics.
Thus, the sentiment relation between $t_c$ and $t_h$ can also be reflected by their semantic similarity. 
If the relation between $t_c$ and $t_h$ is entailment, the semantic distance (the opposite of similarity) between their representations should be as small as possible; conversely, the gap should be as large as possible if their relation is contradiction.
Therefore, we propose to further incorporate the semantic relations between $t_c$ and $t_h$ pairs into modeling their semantic representations.
We first obtain the embeddings of $t_c$ and $t_h$ through the BERT encoder: 
\begin{equation}
\mathbf{z} = \mbox{BERT}(t_c) \in \mathbb R^d,
\end{equation}
\begin{equation}
\boldsymbol{\psi} = \mbox{BERT}(t_h) \in \mathbb R^d,
\end{equation}
where $\mathbf{z}$ and $\boldsymbol{\psi}$ are $d$-dimensional output vectors corresponding to the two special tokens $[\mbox{CLS}]$ as in the following input sequences:[$[\mbox{CLS}], t_c$], [$[\mbox{CLS}], t_h$], respectively.
Next, we calculate their \textit{cosine distance} to represent the semantic similarity:
\begin{equation}
\mbox{distance}(t_c, t_h)=\left(1-\left(\frac{\mathbf{z} \cdot \boldsymbol{\psi}}{\|\mathbf{z}\|\|\boldsymbol{\psi}\|}\right)\right).
\end{equation}
With the goal of narrowing the distance between 
entailment pairs $(t_c,t_h)$, and enlarging the distance between contradiction pairs in the representation space, we design the distance loss function as follows:
\begin{equation}
\mathcal{L}_{\mbox{dist}}=\frac{1}{|T_O|}\sum_{i=1}^{T_O} \mathbb{I}_{(t_c,t_h)_i} \mbox{distance}((t_c,  t_h)_i),
\label{distance_loss}
\end{equation}
where $\mathbb{I}_{(t_c,t_h)_i} \in \{-1,1,0\}$ is an indicator function. The values correspond to the \textit{contradiction} (-1), \textit{entailment} (1), and \textit{neutral} (0) semantic relation between $t_c$ and $t_h$, respectively.

\subsubsection{Sentiment Classification.} The sentiment classification module focuses on predicting the sentiment category of a post.
We still leverage BERT to learn the contextualized representations of posts. Instead of encoding the pair sentence ($t_c$, $t_{h}$), we encode the whole post $t$:

\begin{equation}
\mathbf{z}^c = \mbox{BERT}(t) \in \mathbb R^d,
\label{text_encoder}
\end{equation}
where $\mathbf{z}^c$ is a $d$-dimensional output vector corresponding to the special token $[\mbox{CLS}]$ as in the following input format:[$[\mbox{CLS}], t$].
We pass the learned post vectors through a softmax layer to obtain the probability distribution of M classes:
\begin{equation}
\hat{\mathbf{c}} = p(c|t) = \mbox{softmax}(W^c z^c) \in \mathbb R^M,
\end{equation}
where $W^c \in R^{M\times d}$ is the sentiment classification module's parameters.
We train the model using a cross-entropy loss with the ground-truth sentiment labels of the training data: 
\begin{equation}
\mathcal{L}_{\mbox{sent}}=-\frac{1}{T_O}\sum_{i=1}^{T_O} \sum_{j=1}^{M} c_{i j} \log p\left(c_{ij} \mid {t_i}\right),
\end{equation}
where $c_{i j}$ denotes the ground truth label of $i$-th sample on $j$-th class.
The sentiment classification module is expected to capture important sentiment features and learn to discriminate the sentiment category. At prediction time, we predict the class of post $t_i$ by selecting the largest value of $p(c|t)$ and output $\argmax_j p(c_{ij}|t_i)$. 

\subsubsection{Training and Optimization.} Instead of training the above modules individually, we propose to learn these two tasks simultaneously. We think the learned knowledge from one task can benefit the other in a multi-task learning framework. Given a post that contains sentiment hashtags, we take input $t_i$ and its corresponding pair $(t_c, t_h)_i$ to predict the semantic relation of $(t_c, t_h)_i$ and the sentiment category of $t_i$ simultaneously. 
The proposed \ours~consists of three encoders: a lexicon encoder, a feature encoder, and a task encoder. 
The lexicon encoder converts the input sentence (or sentence pair) into a sequence of token representation vectors, constructed by summing the token, positional, and segment embeddings.
The feature encoder is used to learn contextual word embedding vectors by capturing local context information. 
The task encoder learns task-specific supportive features. 
In \ours, the lexicon encoder and feature extractor layers are shared but with different task-specific output layers. Simultaneously training the semantic inference and sentiment classification modules will allow the feature encoder in \ours~to share fine-grained sentiment features, and thus can better predict the sentiment categories of posts.
Especially when the sentiment signals in posts and hashtags are inconsistent, the sentiment inference module can mitigate such controversy.

We design and combine two objectives for the training of the proposed framework: (1) inference on the semantic relations between posts and hashtags, and (2) predict sentiment categories of posts. Therefore, the overall training loss of \ours~is defined as:
\begin{equation}
\mathcal{L}=\alpha \cdot \mathcal{L}_{\mbox{sent}}+\beta \cdot \mathcal{L}_{\mbox{infer}} + \gamma \mathcal{L}_{\mbox{dist}} ,
\end{equation}
where $\alpha$, $\beta$, $\gamma$ are hyperparameters to control the weights of sentiment classification loss, semantic relation inference loss, and the distance loss.

\subsubsection{Data Augmentation.} 

\begin{figure*}[t]
\centering
\includegraphics[width=1.0\textwidth]{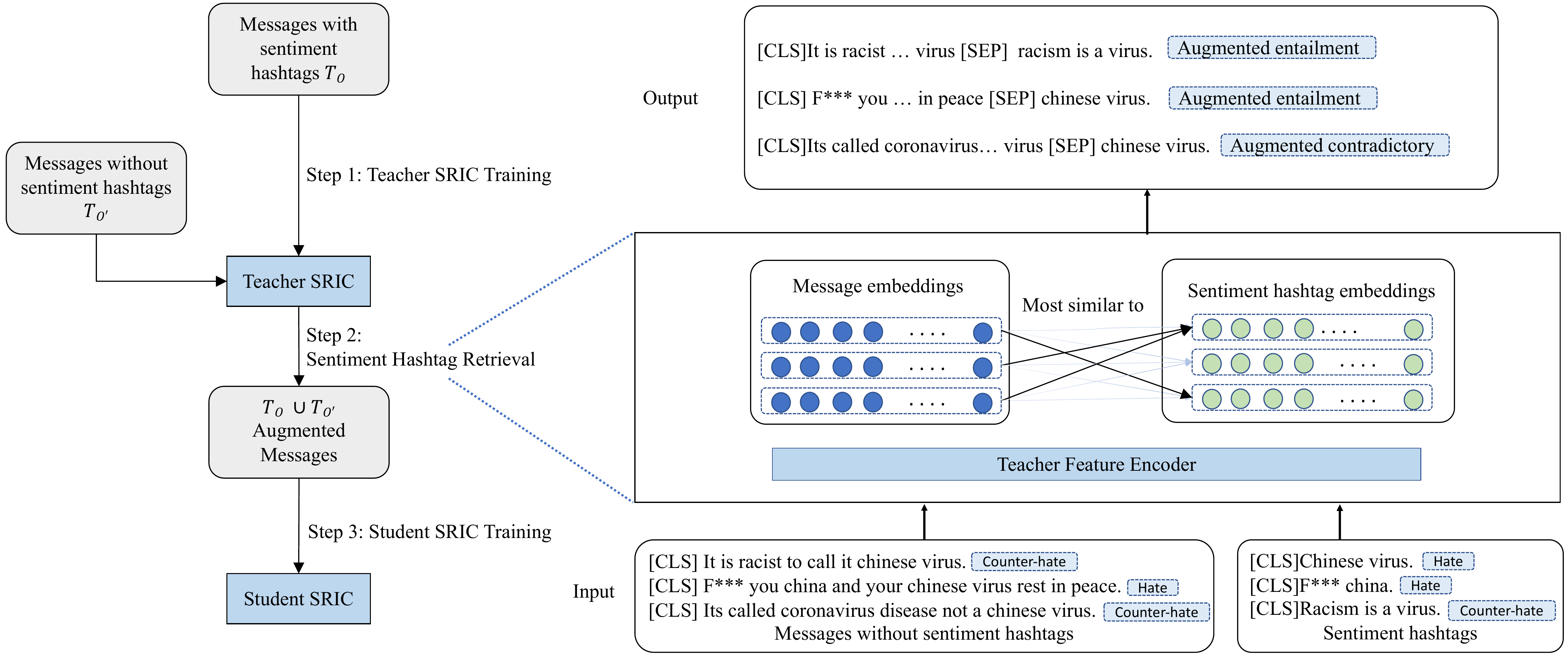} 
\caption{Data augmentation approach. The framework consists of two models, where the teacher SRIC is trained using the dataset that contains sentiment hashtags and the student SRIC is trained using the complete dataset after applying data augmentation method on the dataset without sentiment hashtags. }
\label{infer_aug}
\end{figure*}

We have introduced how to train \ours~using posts that contain sentiment hashtags ($T_O$). However, in reality, there are labeled data ($T_{O'}$) without any sentiment hashtags. In order to utilize these data samples for training, we propose a data augmentation method. 
The method aims to create more data for semantic relation inference learning by matching a potential sentiment hashtag to the posts in $T_{O'}$.
The framework is shown in Figure \ref{infer_aug}.
In the first step, we train a teacher model of \ours~using the $T_O$ dataset.
In the second step, we retrieve the potential sentiment hashtags for posts in $T_{O'}$. 
We first apply the feature encoder in the teacher \ours~model to obtain the embeddings of all segmented hashtags and the embeddings of posts in the $T_{O'}$ dataset: 

\begin{equation}
\boldsymbol{\psi}_{k} = \mbox{BERT}_{\mbox{teacher}}(h_k) \in \mathbb R^d,
\end{equation}
\begin{equation}
\mathbf{z}_{i'} = \mbox{BERT}_{\mbox{teacher}}(t_{i'}) \in \mathbb R^d,
\end{equation}
where the input text format is identical to the one in Equation~\ref{text_encoder}, $h_k \in H$ and $t_{i'} \in T_{O'}$.

Next, for each post, we calculate the similarity score of $z_{i'}$ with all hashtag representation vectors. We choose the hashtag with the largest similarity score as the potential pseudo-hashtag for posts $t_{i'}$:

\begin{equation}
\hat{h}_{k, i'} =\underset{k \in K}{\arg\,\max\,} \mbox{similarity}\left(\mathbf{z_{i'}}, \boldsymbol{\psi}_{k}\right).
\end{equation}

Afterwards, the pseudo semantic relation between the selected hashtag and the post can be determined according to the principle illustrated in Table \ref{relation}. 
For example, if a counter-hate post is matched with a hateful (or counter-hate) hashtag, their semantic relation is contradiction (or entailment).
In the final step, the augmented dataset $T_{O'}$ and $T_O$ are used to train a student \ours.

The similarity measure may create three types of augmented sample pairs with the semantic relation: \textit{entailment}, \textit{contradiction}, or \textit{neutral}.
The contradiction relation indicates that posts and augmented sentiment hashtags have opposite emotions even though they are semantically similar. This means that the sentiment of such posts is more likely to be misclassified.  
Through the semantic relation inference learning of \ours~on the augmented dataset, we expect to further enhance alignment between entailment augmented pairs and improve discrimination between contradiction pairs. 
It is non-trivial to use a similarity score to select potential sentiment hashtags for posts. It serves as an effective way to identify \emph{hard} samples~\cite{DBLP:journals/corr/abs-2104-08821,DBLP:conf/emnlp/KarpukhinOMLWEC20} for training the model to learn to distinguish them.
There are two cases where posts are tended to be misclassified. 
First, a hateful emotion may be expressed implicitly or sarcastically.
Second, counter-hate posts may quote a hateful word (or phrase) to manifest opposition opinion.
In both cases, the model may not be able to capture the real sentiment tendency. 
After augmenting sentiment hashtags to these data, the semantic relation inference in \ours~will embed sentiment information into the representation of posts according to pseudo relation labels, and allow the model to learn sentiment indicative representations.

\section{Experiments}

\subsection{Dataset}
We evaluate the proposed method on two public datasets: Anti-Asian Hate dataset~\cite{DBLP:journals/corr/abs-2005-12423} and  East Asian Prejudice dataset~\cite{DBLP:conf/acl-alw/VidgenHGMBWBHT20}.
The Anti-Asian Hate dataset contains 2,358 labeled tweets where 808 tweets have sentiment hashtags and 1550 tweets do not.
The East Asian Prejudice dataset contains 20,000 labeled tweets. This dataset includes many tweets that do not directly relate to either COVID-19 or East Asia. In our experiments, we remove these tweets and end up keeping 4916 related tweets where 3860 tweets contain sentiment hashtags and 1056 tweets do not.
The label distributions of these two datasets are shown in Table~\ref{data}. Before conducting experiments, we remove hyperlinks, numbers, usernames (i.e., terms that start with `@'), emojis, and punctuation in tweets. We think this information is not helpful for understanding the semantics of tweets. 
\begin{table}[t]
\centering
\begin{tabular}{lccc}
\toprule
    Dataset & Hate & CounterHate & Neutral  \\
\midrule
    Anti-Asian Hate & 811& 362 &1186\\
    East Asian Prejudice& 3931 &89 &896 \\
\bottomrule
\end{tabular}
\caption{Dataset Statistics}
\label{data}
\end{table}

We apply 5-fold cross-validation on the datasets to get the training and test data, where 20\% of the training data is used as the validation data. The validation data is used to determine the number of iterations. We apply early stopping in the training process when the validation loss fails to improve for 5 epochs to avoid overfitting.

\subsection{Baselines}
Sentence representation is a crucial component for semantic-based sentiment classification. Therefore, we compare our proposed model with several state-of-the-art methods listed as below.
\begin{itemize}
\item \textbf{Logistic Regression (LR)}. LR has been widely used as a baseline classification algorithm in NLP. We vectorize tweets using TF-IDF features and then pass the vectors to a LR model to get predictions. We investigate whether the characteristics of word distributions are effective in detecting hate speech. 
\item \textbf{Long short-term memory (LSTM)}~\cite{hochreiter1997long}. As a variant of Recurrent Neural Networks (RNN), LSTM has been proved to have a better ability to learn long-range dependencies in text. In our experiments, we implement an LSTM model with 32 hidden units for one LSTM layer. Besides, we introduce a 20\% dropout and L2 regularization to alleviate overfitting. Finally, the hidden features are passed to a dense layer with a softmax function to get predictions.   
\item \textbf{Convolutional Neural Networks for Sentence Classification (TextCNN)}~\cite{DBLP:journals/corr/Kim14f}. TextCNN has proved to be very successful when it comes to word-level or character-level sentence classification. It is capable of capturing n-gram features. In our experiments, TextCNN model is implemented with three convolution layers with the kernel size from \{3, 4, 5\}, and 128 filters for each layer with the Relu activation function. Afterwards, feature maps are passed to a maxpooling layer following by a concatenation operation to output a hidden vector. We apply a 20\% dropout. The final vector is passed to a dense layer with the softmax function to get predictions. 
\item \textbf{Bidirectional Encoder Representations from Transformers (BERT)}~\cite{DBLP:conf/naacl/DevlinCLT19}. BERT has been proved to be powerful to learn contextulized word and sentence representations. 
We utilize the pre-trained bert-small model as a feature extractor, and take its pooled output as the representations of posts. We also apply a 20\% dropout. The post vectors are then passed to a dense layer with a softmax function to get predictions. 

\end{itemize}

\subsection{Hyper-parameter Setting}
The parameters of the models are selected according to their performance on the validation data. 
For the LSTM model, we tune the hidden dimensions from \{32, 64, 128\}. For the TextCNN model, the window size is adapted from the original paper~\cite{DBLP:journals/corr/Kim14f} and the number of feature maps is chosen from \{64, 128\}. The regularization weight and dropout rate are chosen from  \{0.001, 0.01, 0.1\} and \{0.2, 0.3, 0.4\}, respectively. For the \ours~ model, the loss weights $\alpha$, $\beta$ and $\gamma$ are chosen from 1 to 5.
The LSTM and TextCNN models use Glove pretrained word embeddings of dimension $100$ to initialize inputs and are trained with the Adam optimizer with a learning rate of 1e-4.
Bert and SRIC models are trained with the AdamW optimizer with a learning rate of 3e-5. 
The experiment results are reported as the average scores across 5 different test sets.

\subsection{Evaluation}
Because the labels in the datasets are imbalanced, we adopt weighted $F_1$ score, weighted precision score, weighted recall score, and accuracy (ACC) as evaluation metrics. These metrics are also commonly used in hate speech detection. Here ``weighted'' means that metrics are calculated for each label, and are reported as the weighted average by the number of true instances for each label. For example, assume $y$ is a set of ground-truth labels and $\hat{y}$ is a set of predicted labels. The weighted recall score is calculated as follows:
\begin{equation}
\mbox{weighted recall} = \frac{1}{N} \sum_{m \in M}n_{m} \times s ,
\end{equation}
where $N$ is the total number of training examples, $M$ is the number of pre-defined sentiment labels, $n_m$ is the number of training examples in category $m$, and $s$ is the recall score for category $m$.

\subsection{Experimental Setup}
We conduct two sets of experiments. In the first set of experiments, we attempt to demonstrate the effectiveness of proposed \ours. The models are trained and tested on tweets that contain sentiment hashtags ($T_O$), where 5-fold cross-validation is applied to get the training and test data.  In the second set of experiments, tweets without sentiment hashtags ($T_{O'}$) are added to the training data from the first experiment, while the test data remains unchanged. 
For baseline models, $T_{O'}$ is directly added to the training data. For \ours, we first apply the proposed \textbf{data augmentation method} to retrieve potential sentiment hashtags for tweets in $T_{O'}$, and then the augmented $T_{O'}$ is added to the training data. 
Both sets of experiments are conducted under three configurations: 1) tweets without sentiment hashtags, 2) tweets with sentiment hashtags as single tokens, and 3) tweets with segmented sentiment hashtags. 
We attempt to investigate whether the sentiment-related information embedded in hashtags is helpful for detecting hate speech.

\subsection{Results}
\begin{table*}[t]
\centering
\begin{tabular}{clcccc}
\toprule
    \textbf{Mode} & \textbf{Model} & \textbf{ACC} & \textbf{F1} & \textbf{Precision} & \textbf{Recall} \\
\midrule
    \multirow{4}{*}{\shortstack{without sentiment hashtags}} &LR & 0.6712 (0.036) & 0.6471 (0.039) & 0.6891 (0.029)& 0.6712 (0.036) \\
    & LSTM & 0.6465 (0.017)& 0.6235 (0.032)& 0.6603 (0.035)& 0.6465 (0.017)\\
    & TextCNN & 0.6526 (0.036) & 0.6424 (0.038) & 0.6525 (0.038)& 0.6526 (0.036) \\
    & BERT & 0.6723 (0.042)& 0.6658 (0.043) & 0.6717 (0.041)& 0.6723 (0.020) \\
\midrule
    \multirow{4}{*}{\shortstack{sentiment hashtags\\ (e.g., \#chinavirus)}} &LR & 0.6835 (0.024) & 0.6633 (0.024) & 0.6879 (0.018)& 0.6835 (0.024) \\
    & LSTM & 0.6650 (0.034)& 0.6488 (0.038)& 0.6664 (0.028)& 0.6650 (0.034)\\
    & TextCNN & 0.6551 (0.026) & 0.6465 (0.029) & 0.6568 (0.038)& 0.6551 (0.026) \\
    & BERT & 0.6786 (0.020)& 0.6719 (0.022) & 0.6852 (0.017)& 0.6786 (0.020) \\
\midrule
    
   \multirow{4}{*}{\shortstack{segmented sentiment \\hashtags (e.g., china virus)}} & LR & 0.6922 (0.018)& 0.6713 (0.019) & 0.6922 (0.014)& 0.6922 (0.018) \\
    & LSTM & 0.6725 (0.010) & 0.6610 (0.010) & 0.6741 (0.012) & 0.6725 (0.010) \\
    & TextCNN & 0.6826 (0.028)& 0.6739 (0.027) & 0.6818 (0.028) & 0.6822 (0.028) \\
    & BERT & 0.6885 (0.016)  & 0.6861 (0.023) & 0.6984 (0.022) & 0.6885 (0.016) \\
\midrule
\multirow{1}{*}{\shortstack{inference learning}} & \textbf{\ours~} & 0.7120 (0.024)& 0.7094 (0.026) & 0.7171 (0.015) & 0.7120 (0.024) \\
\bottomrule
\end{tabular}
\caption{Sentiment classification results on tweets with sentiment hashtags of Anti-Asian Hate Dataset ($T_O$) }
\label{SRIC_Anti_Asian}
\end{table*}
\begin{table*}[t]
\centering
\begin{tabular}{c l cccc}
\toprule
    \textbf{Mode} & \textbf{Model} & \textbf{ACC} & \textbf{F1} & \textbf{Precision} & \textbf{Recall} \\
\midrule
    \multirow{4}{*}{\shortstack{without sentiment hashtags}} &LR & 0.6551 (0.023) & 0.6314 (0.030) & 0.6720 (0.023)& 0.6551 (0.025) \\
    & LSTM & 0.6563 (0.016)& 0.6325 (0.029)& 0.6655 (0.031)& 0.6563 (0.016)\\
    & TextCNN & 0.6761 (0.021) & 0.6696 (0.020) & 0.6849 (0.021)& 0.6761 (0.021) \\
    & BERT & 0.6909 (0.020)& 0.6871 (0.021) & 0.6916 (0.017)& 0.6909 (0.020) \\
\midrule
   \multirow{4}{*}{\shortstack{sentiment hashtags\\ (e.g., \#chinavirus)}} & LR & 0.6984 (0.014) & 0.6833 (0.020) & 0.7058 (0.014) & 0.6984 (0.020)\\
    & LSTM & 0.6749 (0.036)& 0.6707 (0.027)& 0.6825 (0.031) & 0.6749 (0.036) \\
    & TextCNN & 0.6675 (0.024) & 0.6617 (0.028) & 0.6723 (0.031) & 0.6675 (0.024) \\
    & BERT & 0.7033 (0.010)& 0.6999 (0.012) & 0.7033 (0.010) & 0.7049 (0.010) \\
\midrule
    \multirow{4}{*}{\shortstack{segmented sentiment \\hashtags (e.g., china virus)}} &LR & 0.7157 (0.022) & 0.7069 (0.023) & 0.7142 (0.022) & 0.7157 (0.022) \\
    & LSTM & 0.6972 (0.024) & 0.6952 (0.023)& 0.6992 (0.024) & 0.6972 (0.024)\\
    & TextCNN & 0.7070 (0.021)& 0.7063 (0.024) & 0.7133 (0.026) & 0.7070 (0.021) \\
    & BERT & 0.7205 (0.019) & 0.7184 (0.017) & 0.7310 (0.017) & 0.7205 (0.019) \\
\midrule
    \multirow{1}{*}{\shortstack{inference learning}} &\textbf{\ours~} & 0.7435 (0.026) & 0.7464 (0.026) & 0.7543 (0.027) & 0.7435 (0.026) \\
\bottomrule
\end{tabular}
\caption{Sentiment classification results using augmented training data of Anti-Asian Hate Dataset ($T_O \cup T_{O'}$)}
\label{SRIC_Anti_Asian_Aug}
\end{table*}

\begin{table*}[t]
\centering
\begin{tabular}{clcccc}
\toprule
    \textbf{Mode} & \textbf{Model} & \textbf{ACC} & \textbf{F1} & \textbf{Precision} & \textbf{Recall} \\
\midrule
    \multirow{4}{*}{\shortstack{without sentiment hashtags}} &LR & 0.8746 (0.011) & 0.8578 (0.014) & 0.8531 (0.014)& 0.8746 (0.011) \\
    & LSTM & 0.8869 (0.009)& 0.8756 (0.010)& 0.8679 (0.009)& 0.8869 (0.009)\\
    & TextCNN & 0.8913 (0.011) & 0.8802 (0.012) & 0.8714 (0.011)& 0.8913 (0.012) \\
    & BERT & 0.8854 (0.007)& 0.8766 (0.007) & 0.8687 (0.007)& 0.8854 (0.007) \\
\midrule
    \multirow{4}{*}{\shortstack{sentiment hashtags\\ (e.g., \#chinavirus)}} &LR & 0.8795 (0.009) & 0.8639 (0.011) & 0.8581 (0.009)& 0.8795 (0.009) \\
    & LSTM & 0.8931 (0.006)& 0.8837 (0.007)& 0.8749 (0.006)& 0.8931 (0.006)\\
    & TextCNN & 0.8923 (0.009) & 0.8811 (0.010) & 0.8724 (0.009)& 0.8923 (0.010) \\
    & BERT & 0.8928 (0.007)& 0.8825 (0.007) & 0.8741 (0.007)& 0.8928 (0.007) \\
\midrule
   \multirow{4}{*}{\shortstack{segmented sentiment \\hashtags (e.g., china virus)}} & LR & 0.8767 (0.008)& 0.8601 (0.010) & 0.8551 (0.009)& 0.8767 (0.005) \\
    & LSTM & 0.8835 (0.004) & 0.8739 (0.004) & 0.8651 (0.004) & 0.8835 (0.004) \\
    & TextCNN & 0.8941 (0.013)& 0.8831 (0.014) & 0.8743 (0.013) & 0.8941 (0.015) \\
    & BERT & 0.8936 (0.010)  & 0.8847 (0.012) & 0.8773 (0.010) & 0.8936 (0.012) \\
\midrule
\multirow{1}{*}{\shortstack{inference learning}} & \textbf{\ours~} & 0.9064 (0.004)& 0.8979 (0.007) & 0.8899 (0.012) & 0.9064 (0.004) \\
\bottomrule
\end{tabular}
\caption{Sentiment classification results on tweets with sentiment hashtags of East Asian Prejudice Dataset  ($T_O$)}
\label{SRIC_East_Asian_Prejudice}
\end{table*}
\begin{table*}[t]
\centering
\begin{tabular}{c l cccc}
\toprule
    \textbf{Mode} & \textbf{Model} & \textbf{ACC} & \textbf{F1} & \textbf{Precision} & \textbf{Recall} \\
\midrule
    \multirow{4}{*}{\shortstack{without sentiment hashtags}} &LR & 0.8751 (0.010) & 0.8586 (0.012) & 0. 8536(0.010)& 0.8751 (0.014) \\
    & LSTM & 0.8871 (0.005)& 0.8795 (0.004)& 0.8732 (0.003)& 0.8871 (0.005)\\
    & TextCNN & 0.8934 (0.013) & 0.8818 (0.015) & 0.8728 (0.016)& 0.8934 (0.013) \\
    & BERT & 0.8909 (0.006)& 0.8823 (0.008) & 0.8763 (0.008)& 0. 8909(0.006) \\
\midrule
   \multirow{4}{*}{\shortstack{sentiment hashtags\\ (e.g., \#chinavirus)}} & LR & 0.8803 (0.009) & 0.8647 (0.011) & 0.8590 (0.009) & 0.8803 (0.012)\\
    & LSTM & 0.8918 (0.006)& 0.8835 (0.008)& 0.8759 (0.006) & 0.8918 (0.009) \\
    & TextCNN & 0.8941 (0.007) & 0.8828 (0.007) & 0.8780 (0.007) & 0.8941 (0.005) \\
    & BERT & 0.8936 (0.008)& 0.8852 (0.009) & 0.8795 (0.008) & 0.8936 (0.011) \\
\midrule
    \multirow{4}{*}{\shortstack{segmented sentiment \\hashtags (e.g., china virus)}} &LR & 0.8785 (0.012) & 0.8632 (0.013) & 0.8568 (0.015) & 0.8785 (0.015) \\
    & LSTM & 0.8949 (0.010) & 0.8855 (0.010)& 0.8766 (0.010) & 0.8949 (0.010)\\
    & TextCNN & 0.8954 (0.009)& 0.8851 (0.009) & 0.8760 (0.009) & 0.8954 (0.010) \\
    & BERT & 0.8949 (0.012) & 0.8869 (0.011) & 0.8804 (0.012) & 0.8949 (0.010) \\
\midrule
    \multirow{1}{*}{\shortstack{inference learning}} &\textbf{\ours~} & 0.9113 (0.005) & 0.9007 (0.005) & 0.8923 (0.005) & 0.9113 (0.005) \\
\bottomrule
\end{tabular}
\caption{Sentiment classification results using augmented training data of East Asian Prejudice Dataset ($T_O \cup T_{O'}$)}
\label{SRIC_East_Asian_Prejudice_Aug}
\end{table*}

Table~\ref{SRIC_Anti_Asian} and Table~\ref{SRIC_Anti_Asian_Aug} show the experimental results on the Anti-Asian Hate dataset. 
Compared to the LR model that relies on word distribution features, deep learning models have better results after segmenting sentiment hashtags. This demonstrates that the semantics in sentiment hashtags are useful for learning sentiment-indicative representations of posts, and thus improve the ability of classifiers.
The LR model achieves comparable classification performance compared  with the BERT model. This indicates that the characteristics of word distributions are useful in classifying different sentiments in this dataset. We check the coefficients of features in two LR models that are fed with segmented sentiment hashtags and complete sentiment hashtags as single tokens, respectively.
We find that they both put more weights on words, such as \textit{chink}, \textit{f***}, \textit{ccp}, \textit{shit}, \textit{coronavirus}, \textit{wuhan}, \textit{china}, \textit{asians}, \textit{commie}, \textit{communist}, \textit{racism}, \textit{racist}, \textit{stand}, \textit{support}, and \textit{stophate}. 
The appearance of these words is skewed towards a certain sentiment category. Therefore, the LR model can achieve good classification results relying on these words.
Table~\ref{SRIC_East_Asian_Prejudice} and Table~\ref{SRIC_East_Asian_Prejudice_Aug} exhibit the experimental results on the East-Asian Prejudice dataset. When segmenting sentiment hashtags, we observe a marginal improvement in the baseline models, except for the LR model.
We further analyze the reason why the LR model becomes worse. 
We check the number of features and the feature importance in the decision function. 
With sentiment hashtags as single tokens, the model takes 8,608 words as features, and puts more weights on words, such as \textit{wuhanvirus}, \textit{wuflu}, and \textit{wuhanflu}. With segmented sentiment hashtags, the model gets 8,602 features, and words that are generally distributed among the corpus, such as \textit{wuhan}, are more weighted. We think it is the slight difference in the number of features, and the decrease of sentiment-indictive tokens (e.g.,\textit{wuhanvirus}) that result in the decrease in performance.

The comparison results of posts without sentiment hashtags and posts with sentiment hashtags indicate that sentiment hashtags can improve hate speech detection performance. However, the improvement is limited and there is a risk of introducing conflict sentiment signals into context as we illustrate in Table~\ref{tab:relation-example}, which inspires the proposed \ours~framework.
The experimental results demonstrate that the proposed \ours~outperforms all the baseline models in two sets of experiments. We think it is attributed to the semantic relation inference learning in the framework. 
By modeling the semantic relation between posts and hashtags, the proposed \ours~will not only capture semantic information but also sentiment information, and thus produce more suitable sentiment-indictive features for hate speech detection. 
The additional sentiment features, e.g., sentiment hashtags, are served as prompts to encode sentiment signals into the representation of texts under the guidance of semantic relations.

Comparing the results of two datasets, we have two discoveries.
We first observe that the semantics in sentiment hashtags have more effect on the Anti-Asain dataset than the East Asain Prejudice dataset in the task of sentiment classification. For deep learning baseline models, the overall classification performance on the Anti-Asain dataset is obviously improved, while the improvement on the East Asain Prejudice dataset is not significant with segmented sentiment hashtags.
After analyzing the average length of tweets in two datasets, we find that there are an average 151 words per tweet in the Anti-Asian dataset, and an average of 188 words per tweet in the East Asian Prejudice dataset. We think the sentiment signals brought by sentiment hashtags may be mitigated by their longer sentiment-unrelated context. 
We also notice that proposed \ours~bring more improvement on the Anti-Asian dataset than the East Asian Prejudice dataset. 
There are two groups of sentiment hashtags in the Anti-Asian dataset, hateful hashtags and counter-hate hashtags. The significant difference in their semantics (shown in Figure~\ref{simliarity}) can enable \ours~to learn discriminative representations. However, the East Asian Prejudice data only contains hateful hashtags. 

We further check the number of pseudo semantic relations generated by the proposed \ours~in the data augmentation approach. 
In the Anti-Asian Hate dataset, there are a total of 1,550 samples without sentiment hashtags, among which 586 entailment, 172 contradiction, and 792 neutral relations, are inferred from the semantic similarity measure respectively. 
As the results show that there are more entailment semantic relations generated than contradictions. 
We think it is own to the distance loss (Equation~\ref{distance_loss}) in ~\ours.
There are some posts, however, that contain complex sentimental contexts, resulting in contradiction relations. This is because semantic similarities between posts and augmented sentiment hashtags are partially matched. Taking the post ``reminder calling it the chinese virus is not racist its truth'' as an example, a counter-hate hashtag ``\#stopthehate'' will be matched according to semantic similarity but their sentiment intentions are opposed.
Since the East Asian Prejudice dataset only contains hateful hashtags, the inferred semantic relations by \ours~have the same distribution as posts' sentiment labels.

\section{Conclusion}
In this paper, we propose \ours, a hate speech classification framework with semantic relation inference as enhancement. The semantic relation inference module is designed to learn fine-grained sentiment indicative features under the guidance of sentiment hashtags, while the sentiment classification module aims to capture discriminative features to classify different sentiment categories. We demonstrate the effective classification performance of the proposed framework on two challenging datasets with multi-sentiment categories. 
In the future, we plan to explore two directions: 1) leveraging massive unlabeled corpora to further improve hate speech classifiers, and 2) improving the model's generalization ability on datasets with different sentiment categories and distribution characteristics by providing different hashtags. In our work, we leverage sentiment hashtags in datasets as sentiment descriptions to train the proposed framework. In more general cases, datasets may not contain specific notations that sufficiently indicate emotions. Artificially defined sentiment descriptions based on human knowledge will be considered to apply to such datasets. 
Moreover, we plan to investigate the efficacy of sentimental semantic relation inference under the guidance of sentiment prompts on other sentiment classification tasks.
\section*{Acknowledgements}
This work is supported in part by the US National Science Foundation under grants 1948432 and 2047843. Any opinions, findings, and conclusions or recommendations expressed in this material are those of the authors and do not necessarily reflect the views of the National Science Foundation.

\begin{small}
\bibliography{ref.bib}

\begin{thebibliography}{49}
\providecommand{\natexlab}[1]{#1}

\bibitem[{Badjatiya et~al.(2017)Badjatiya, Gupta, Gupta, and
  Varma}]{DBLP:conf/www/BadjatiyaG0V17}
Badjatiya, P.; Gupta, S.; Gupta, M.; and Varma, V. 2017.
\newblock Deep Learning for Hate Speech Detection in Tweets.
\newblock In \emph{WWW}, 759--760. {ACM}.

\bibitem[{Baruah, Barbhuiya, and Dey(2019)}]{DBLP:conf/semeval/BaruahBD19}
Baruah, A.; Barbhuiya, F.~A.; and Dey, K. 2019.
\newblock {ABARUAH} at SemEval-2019 Task 5 : Bi-directional {LSTM} for Hate
  Speech Detection.
\newblock In \emph{Proceedings of the 13th International Workshop on Semantic
  Evaluation, SemEval@NAACL-HLT}, 371--376. ACL.

\bibitem[{Burnap and Williams(2015)}]{burnap2015cyber}
Burnap, P.; and Williams, M.~L. 2015.
\newblock Cyber hate speech on twitter: An application of machine
  classification and statistical modeling for policy and decision making.
\newblock \emph{Policy \& internet}, 7(2): 223--242.

\bibitem[{Capozzi et~al.(2019)Capozzi, Lai, Basile, Poletto, Sanguinetti,
  Bosco, Patti, Ruffo, Musto, Polignano et~al.}]{capozzi2019computational}
Capozzi, A.~T.; Lai, M.; Basile, V.; Poletto, F.; Sanguinetti, M.; Bosco, C.;
  Patti, V.; Ruffo, G.; Musto, C.; Polignano, M.; et~al. 2019.
\newblock Computational linguistics against hate: Hate speech detection and
  visualization on social media in the" Contro L’Odio" project.
\newblock In \emph{CLiC-it}, volume 2481, 1--6. CEUR-WS.

\bibitem[{Cowan et~al.(2005)Cowan, Heiple, Marquez, Khatchadourian, and
  McNevin}]{cowan2005heterosexuals}
Cowan, G.; Heiple, B.; Marquez, C.; Khatchadourian, D.; and McNevin, M. 2005.
\newblock Heterosexuals' attitudes toward hate crimes and hate speech against
  gays and lesbians: Old-fashioned and modern heterosexism.
\newblock \emph{Journal of Homosexuality}, 49(2): 67--82.

\bibitem[{Davidov, Tsur, and Rappoport(2010)}]{davidov2010enhanced}
Davidov, D.; Tsur, O.; and Rappoport, A. 2010.
\newblock Enhanced sentiment learning using twitter hashtags and smileys.
\newblock In \emph{Coling:Posters}, 241--249.

\bibitem[{Davidson, Bhattacharya, and Weber(2019)}]{davidson2019racial}
Davidson, T.; Bhattacharya, D.; and Weber, I. 2019.
\newblock Racial Bias in Hate Speech and Abusive Language Detection Datasets.
\newblock \emph{CoRR}, abs/1905.12516.

\bibitem[{Davidson et~al.(2017)Davidson, Warmsley, Macy, and
  Weber}]{DBLP:conf/icwsm/DavidsonWMW17}
Davidson, T.; Warmsley, D.; Macy, M.~W.; and Weber, I. 2017.
\newblock Automated Hate Speech Detection and the Problem of Offensive
  Language.
\newblock In \emph{ICWSM}, 512--515. {AAAI} Press.

\bibitem[{Devlin et~al.(2019)Devlin, Chang, Lee, and
  Toutanova}]{DBLP:conf/naacl/DevlinCLT19}
Devlin, J.; Chang, M.; Lee, K.; and Toutanova, K. 2019.
\newblock {BERT:} Pre-training of Deep Bidirectional Transformers for Language
  Understanding.
\newblock In \emph{{NAACL-HLT}}, 4171--4186. Association for Computational
  Linguistics.

\bibitem[{dos Santos and Gatti(2014)}]{DBLP:conf/coling/SantosG14}
dos Santos, C.~N.; and Gatti, M. 2014.
\newblock Deep Convolutional Neural Networks for Sentiment Analysis of Short
  Texts.
\newblock In \emph{COLING}, 69--78. {ACL}.

\bibitem[{Englmeier and Mothe(2020)}]{DBLP:conf/ahfe/EnglmeierM20}
Englmeier, K.; and Mothe, J. 2020.
\newblock Application-Oriented Approach for Detecting Cyberaggression in Social
  Media.
\newblock In \emph{AHFE}, volume 1213 of \emph{Advances in Intelligent Systems
  and Computing}, 129--136. Springer.

\bibitem[{Fan, Yu, and Yin(2020)}]{fan2020stigmatization}
Fan, L.; Yu, H.; and Yin, Z. 2020.
\newblock Stigmatization in social media: Documenting and analyzing hate speech
  for COVID-19 on Twitter.
\newblock In \emph{Proceedings of the Association for Information Science and
  Technology}, volume~57, e313. Wiley Online Library.

\bibitem[{Fortuna and Nunes(2018)}]{fortuna2018survey}
Fortuna, P.; and Nunes, S. 2018.
\newblock A survey on automatic detection of hate speech in text.
\newblock \emph{ACM Computing Surveys}, 51(4): 1--30.

\bibitem[{Fr{\'\i}as-V{\'a}zquez and Arcila(2019)}]{frias2019hate}
Fr{\'\i}as-V{\'a}zquez, M.; and Arcila, C. 2019.
\newblock Hate speech against Central American immigrants in Mexico: Analysis
  of xenophobia and racism in politicians, media and citizens.
\newblock In \emph{TEEM}, 956--960. {ACM}.

\bibitem[{Fr{\'{\i}}as{-}V{\'{a}}zquez and P{\'{e}}rez(2019)}]{vazquez2019hate}
Fr{\'{\i}}as{-}V{\'{a}}zquez, M.; and P{\'{e}}rez, F.~S. 2019.
\newblock Hate Speech in Spain Against Aquarius Refugees 2018 in Twitter.
\newblock In \emph{TEEM}, 906--910. {ACM}.

\bibitem[{Gamb{\"a}ck and Sikdar(2017)}]{Gambck2017UsingCN}
Gamb{\"a}ck, B.; and Sikdar, U.~K. 2017.
\newblock Using Convolutional Neural Networks to Classify Hate-Speech.
\newblock In \emph{ALW@ACL}.

\bibitem[{Gao, Yao, and Chen(2021)}]{DBLP:journals/corr/abs-2104-08821}
Gao, T.; Yao, X.; and Chen, D. 2021.
\newblock SimCSE: Simple Contrastive Learning of Sentence Embeddings.
\newblock \emph{CoRR}, abs/2104.08821.

\bibitem[{Georgakopoulos et~al.(2018)Georgakopoulos, Tasoulis, Vrahatis, and
  Plagianakos}]{DBLP:conf/setn/GeorgakopoulosT18}
Georgakopoulos, S.~V.; Tasoulis, S.~K.; Vrahatis, A.~G.; and Plagianakos, V.~P.
  2018.
\newblock Convolutional Neural Networks for Toxic Comment Classification.
\newblock In \emph{SETN}, 35:1--35:6. {ACM}.

\bibitem[{Grimminger and Klinger(2021)}]{DBLP:conf/wassa/GrimmingerK21}
Grimminger, L.; and Klinger, R. 2021.
\newblock Hate Towards the Political Opponent: {A} Twitter Corpus Study of the
  2020 {US} Elections on the Basis of Offensive Speech and Stance Detection.
\newblock In \emph{Proceedings of the Eleventh Workshop on Computational
  Approaches to Subjectivity, Sentiment and Social Media Analysis}, 171--180.
  ACL.

\bibitem[{Hardage and Najafirad(2020)}]{DBLP:conf/emnlp/HardageN20}
Hardage, D.; and Najafirad, P. 2020.
\newblock Hate and Toxic Speech Detection in the Context of Covid-19 Pandemic
  using {XAI:} Ongoing Applied Research.
\newblock In \emph{Proceedings of the 1st Workshop on {NLP} for
  COVID-19@{EMNLP}}. ACL.

\bibitem[{Hochreiter and Schmidhuber(1997)}]{hochreiter1997long}
Hochreiter, S.; and Schmidhuber, J. 1997.
\newblock Long short-term memory.
\newblock \emph{Neural computation}, 9(8): 1735--1780.

\bibitem[{Indurthi et~al.(2019)Indurthi, Syed, Shrivastava, Chakravartula,
  Gupta, and Varma}]{indurthi2019fermi}
Indurthi, V.; Syed, B.; Shrivastava, M.; Chakravartula, N.; Gupta, M.; and
  Varma, V. 2019.
\newblock Fermi at semeval-2019 task 5: Using sentence embeddings to identify
  hate speech against immigrants and women in twitter.
\newblock In \emph{NAACL-HLT}, 70--74. ACL.

\bibitem[{Jiang and de~Marneffe(2019)}]{DBLP:conf/emnlp/JiangM19}
Jiang, N.; and de~Marneffe, M. 2019.
\newblock Evaluating {BERT} for natural language inference: {A} case study on
  the CommitmentBank.
\newblock In Inui, K.; Jiang, J.; Ng, V.; and Wan, X., eds.,
  \emph{{EMNLP-IJCNLP}}, 6085--6090. ACL.

\bibitem[{Karpukhin et~al.(2020)Karpukhin, Oguz, Min, Lewis, Wu, Edunov, Chen,
  and Yih}]{DBLP:conf/emnlp/KarpukhinOMLWEC20}
Karpukhin, V.; Oguz, B.; Min, S.; Lewis, P. S.~H.; Wu, L.; Edunov, S.; Chen,
  D.; and Yih, W. 2020.
\newblock Dense Passage Retrieval for Open-Domain Question Answering.
\newblock In Webber, B.; Cohn, T.; He, Y.; and Liu, Y., eds., \emph{{EMNLP}},
  6769--6781. ACL.

\bibitem[{Kim(2014)}]{DBLP:journals/corr/Kim14f}
Kim, Y. 2014.
\newblock Convolutional Neural Networks for Sentence Classification.
\newblock \emph{CoRR}, abs/1408.5882.

\bibitem[{Koto and Adriani(2015)}]{koto2015hbe}
Koto, F.; and Adriani, M. 2015.
\newblock HBE: Hashtag-based emotion lexicons for twitter sentiment analysis.
\newblock In \emph{FIRE}, 31--34.

\bibitem[{Kouloumpis, Wilson, and Moore(2011)}]{kouloumpis2011twitter}
Kouloumpis, E.; Wilson, T.; and Moore, J. 2011.
\newblock Twitter sentiment analysis: The good the bad and the omg!
\newblock In \emph{ICWSM}.

\bibitem[{Kwok and Wang(2013)}]{kwok2013locate}
Kwok, I.; and Wang, Y. 2013.
\newblock Locate the Hate: Detecting Tweets against Blacks.
\newblock In \emph{AAAI}. {AAAI} Press.

\bibitem[{Maddela, Xu, and Preotiuc{-}Pietro(2019)}]{maddela2019multi}
Maddela, M.; Xu, W.; and Preotiuc{-}Pietro, D. 2019.
\newblock Multi-task Pairwise Neural Ranking for Hashtag Segmentation.
\newblock In \emph{ACL}, 2538--2549.

\bibitem[{Melton, Bagavathi, and Krishnan(2020)}]{DBLP:conf/icmla/MeltonBK20}
Melton, J.; Bagavathi, A.; and Krishnan, S. 2020.
\newblock DeL-haTE: {A} Deep Learning Tunable Ensemble for Hate Speech
  Detection.
\newblock In Wani, M.~A.; Luo, F.; Li, X.~A.; Dou, D.; and Bonchi, F., eds.,
  \emph{ICMLA}, 1015--1022. {IEEE}.

\bibitem[{Mozafari, Farahbakhsh, and
  Crespi(2019)}]{DBLP:conf/complexnetworks/MozafariFC19}
Mozafari, M.; Farahbakhsh, R.; and Crespi, N. 2019.
\newblock A BERT-Based Transfer Learning Approach for Hate Speech Detection in
  Online Social Media.
\newblock In \emph{CNA}, volume 881, 928--940. Springer.

\bibitem[{Narang and Brew(2020)}]{DBLP:conf/acl-alw/NarangB20}
Narang, K.; and Brew, C. 2020.
\newblock Abusive Language Detection using Syntactic Dependency Graphs.
\newblock In \emph{Proceedings of the Fourth Workshop on Online Abuse and
  Harms, {WOAH}}, 44--53. ACL.

\bibitem[{Qian et~al.(2021)Qian, Wang, ElSherief, and
  Yan}]{DBLP:conf/naacl/QianWEY21}
Qian, J.; Wang, H.; ElSherief, M.; and Yan, X. 2021.
\newblock Lifelong Learning of Hate Speech Classification on Social Media.
\newblock In \emph{NAACL-HLT}, 2304--2314. ACL.

\bibitem[{Rezapour et~al.(2017)Rezapour, Wang, Abdar, and
  Diesner}]{rezapour2017identifying}
Rezapour, R.; Wang, L.; Abdar, O.; and Diesner, J. 2017.
\newblock Identifying the overlap between election result and candidates’
  ranking based on hashtag-enhanced, lexicon-based sentiment analysis.
\newblock In \emph{ICSC}, 93--96. IEEE.

\bibitem[{Saha et~al.(2018)Saha, Mathew, Goyal, and
  Mukherjee}]{saha2018hateminers}
Saha, P.; Mathew, B.; Goyal, P.; and Mukherjee, A. 2018.
\newblock Hateminers : Detecting Hate speech against Women.
\newblock \emph{CoRR}, abs/1812.06700.

\bibitem[{Sap et~al.(2019)Sap, Card, Gabriel, Choi, and Smith}]{sap2019risk}
Sap, M.; Card, D.; Gabriel, S.; Choi, Y.; and Smith, N.~A. 2019.
\newblock The risk of racial bias in hate speech detection.
\newblock In \emph{ACL}, 1668--1678.

\bibitem[{Schmidt and Wiegand(2017{\natexlab{a}})}]{schmidt2017survey}
Schmidt, A.; and Wiegand, M. 2017{\natexlab{a}}.
\newblock A survey on hate speech detection using natural language processing.
\newblock In \emph{ACL}, 1--10.

\bibitem[{Schmidt and
  Wiegand(2017{\natexlab{b}})}]{DBLP:conf/acl-socialnlp/SchmidtW17}
Schmidt, A.; and Wiegand, M. 2017{\natexlab{b}}.
\newblock A Survey on Hate Speech Detection using Natural Language Processing.
\newblock In \emph{Proceedings of the Fifth International Workshop on Natural
  Language Processing for Social Media, SocialNLP@EACL}, 1--10. ACL.

\bibitem[{Siegel et~al.(2021)Siegel, Nikitin, Barber{\'a}, Sterling, Pullen,
  Bonneau, Nagler, Tucker et~al.}]{siegel2021trumping}
Siegel, A.~A.; Nikitin, E.; Barber{\'a}, P.; Sterling, J.; Pullen, B.; Bonneau,
  R.; Nagler, J.; Tucker, J.~A.; et~al. 2021.
\newblock Trumping Hate on Twitter? Online Hate Speech in the 2016 US Election
  Campaign and its Aftermath.
\newblock \emph{Quarterly Journal of Political Science}, 16(1): 71--104.

\bibitem[{Sohn and Lee(2019)}]{DBLP:conf/icdm/SohnL19}
Sohn, H.; and Lee, H. 2019.
\newblock {MC-BERT4HATE:} Hate Speech Detection using Multi-channel {BERT} for
  Different Languages and Translations.
\newblock In \emph{2019 International Conference on Data Mining Workshops,
  {ICDM} Workshops}, 551--559. {IEEE}.

\bibitem[{Vargas et~al.(2021)Vargas, de~G{\'{o}}es, Carvalho, Benevenuto, and
  Pardo}]{DBLP:journals/corr/abs-2104-12265}
Vargas, F.~A.; de~G{\'{o}}es, F.~R.; Carvalho, I.; Benevenuto, F.; and Pardo,
  T. A.~S. 2021.
\newblock Contextual Lexicon-Based Approach for Hate Speech and Offensive
  Language Detection.
\newblock \emph{CoRR}, abs/2104.12265.

\bibitem[{Vidgen et~al.(2020)Vidgen, Hale, Guest, Margetts, Broniatowski,
  Waseem, Botelho, Hall, and Tromble}]{DBLP:conf/acl-alw/VidgenHGMBWBHT20}
Vidgen, B.; Hale, S.~A.; Guest, E.; Margetts, H.~Z.; Broniatowski, D.~A.;
  Waseem, Z.; Botelho, A.; Hall, M.; and Tromble, R. 2020.
\newblock Detecting East Asian Prejudice on Social Media.
\newblock In \emph{Proceedings of the Fourth Workshop on Online Abuse and
  Harms}, 162--172. ACL.

\bibitem[{Vishwamitra et~al.(2020)Vishwamitra, Hu, Luo, Cheng, Costello, and
  Yang}]{DBLP:conf/icmla/VishwamitraH0CC20}
Vishwamitra, N.; Hu, R.~R.; Luo, F.; Cheng, L.; Costello, M.; and Yang, Y.
  2020.
\newblock On Analyzing COVID-19-related Hate Speech Using {BERT} Attention.
\newblock In \emph{ICMLA}, 669--676. {IEEE}.

\bibitem[{Wang et~al.(2011)Wang, Wei, Liu, Zhou, and Zhang}]{wang2011topic}
Wang, X.; Wei, F.; Liu, X.; Zhou, M.; and Zhang, M. 2011.
\newblock Topic sentiment analysis in twitter: a graph-based hashtag sentiment
  classification approach.
\newblock In \emph{CIKM}, 1031--1040.

\bibitem[{Waseem and Hovy(2016)}]{DBLP:conf/naacl/WaseemH16}
Waseem, Z.; and Hovy, D. 2016.
\newblock Hateful Symbols or Hateful People? Predictive Features for Hate
  Speech Detection on Twitter.
\newblock In \emph{Proceedings of the Student Research Workshop,
  SRW@HLT-NAACL}, 88--93. ACL.

\bibitem[{Xia, Field, and Tsvetkov(2020)}]{xia-etal-2020-demoting}
Xia, M.; Field, A.; and Tsvetkov, Y. 2020.
\newblock Demoting Racial Bias in Hate Speech Detection.
\newblock In \emph{Proceedings of the Eighth International Workshop on Natural
  Language Processing for Social Media}, 7--14. ACL.

\bibitem[{Zhang and Luo(2019)}]{zhang2019hate}
Zhang, Z.; and Luo, L. 2019.
\newblock Hate speech detection: A solved problem? the challenging case of long
  tail on twitter.
\newblock \emph{Semantic Web}, 10(5): 925--945.

\bibitem[{Zhong et~al.(2016)Zhong, Li, Squicciarini, Rajtmajer, Griffin,
  Miller, and Caragea}]{DBLP:conf/ijcai/ZhongLSRG0C16}
Zhong, H.; Li, H.; Squicciarini, A.~C.; Rajtmajer, S.~M.; Griffin, C.; Miller,
  D.~J.; and Caragea, C. 2016.
\newblock Content-Driven Detection of Cyberbullying on the Instagram Social
  Network.
\newblock In Kambhampati, S., ed., \emph{IJCAI}, 3952--3958. {IJCAI/AAAI}
  Press.

\bibitem[{Ziems et~al.(2020)Ziems, He, Soni, and
  Kumar}]{DBLP:journals/corr/abs-2005-12423}
Ziems, C.; He, B.; Soni, S.; and Kumar, S. 2020.
\newblock Racism is a Virus: Anti-Asian Hate and Counterhate in Social Media
  during the {COVID-19} Crisis.
\newblock \emph{CoRR}, abs/2005.12423.

\end{thebibliography}
\end{small}%
\end{document}